\documentclass{article} 
\usepackage{iclr2023_conference_tinypaper,times,graphicx,longtable}

\usepackage{amsmath,amsfonts,bm}

\def\eqref#1{equation~\ref{#1}}

\def\1{\bm{1}}

\DeclareMathAlphabet{\mathsfit}{\encodingdefault}{\sfdefault}{m}{sl}
\SetMathAlphabet{\mathsfit}{bold}{\encodingdefault}{\sfdefault}{bx}{n}

\usepackage{hyperref}
\usepackage{url}

\title{Hard ASH: Sparsity and the right optimizer make a continual learner}

\author{Santtu Keskinen \\
Unaffiliated \\
\texttt{santtu.keskinen@gmail.com}
}

\iclrfinalcopy 
\begin{document}

\maketitle
\vspace{-0.5cm}
\begin{abstract}
In class incremental learning, neural networks typically suffer from catastrophic forgetting. We show that an MLP featuring a sparse activation function and an adaptive learning rate optimizer can compete with established regularization techniques in the Split-MNIST task. We highlight the effectiveness of the Adaptive SwisH (ASH) activation function in this context and introduce a novel variant, Hard Adaptive SwisH (Hard ASH) to further enhance the learning retention.
\end{abstract}

\vspace{-0.2cm}
\section{Introduction}
Continual learning presents a unique challenge for artificial neural networks, particularly in the class incremental setting \citep{hsu2019reevaluating}, where a single network must remember old classes that have left the training set. In this paper I explore an overlooked approach that doesn't require any techniques developed specifically for continual learning. For regularization I used only carefully tuned optimizers with adaptive learning rate such as Adagrad \citep{adagrad}. The approach does not exploit the task structure in any way. This is in contrast to most regularizing continual learning methods that require either explicit task boundaries such as EWC \citep{Kirkpatrick_2017} and MAS \citep{aljundi2018memory} or implied task boundaries like Online EWC \citep{schwarz2018progress}. Perhaps closest to my method are the Elephant MLP \citep{lan2023elephant} and SDMLP \citep{bricken2023sparse}, but the results here outperform both in Split-MNIST with an arguably simpler method.

Sparse representations have been shown to be effective at reducing forgetting in neural networks \citep{srivastava_2013, shen2021algorithmic, ahmad2019dense, lan2023elephant}. I continue this pattern and show that combining sparsity with an adaptive learning rate optimizer is enough to make a conceptually simple but surprisingly effective continual learner.

To make my MLP hidden layer representations sparse, I used an activation function that makes the majority of the activations zero. Top-K (also known as k-WTA) is the conceptually simplest sparse activation function and it's usage in neural networks goes back to at least \citet{makhzani2014ksparse}. I show that Top-K works well in my setup, but I get better accuracy with my novel Hard Adaptive Swish (Hard ASH) activation.

\vspace{-0.1cm}
\section{ASH and Hard ASH}
\vspace{-0.1cm}
The Adaptive SwisH (ASH) activation function \citep{lee2022stochastic}, introduced a new way of controlling the amount of sparsity of the activations, that is cheaper to compute than the Top-K function. This is the first study to use ASH for continual learning. The formulation for ASH I use is:
\vspace{-0.1cm}
\begin{center}
\
$ASH(x_i) = x_i \cdot  S(\alpha \cdot ( x_i - \mu_X - z_k \cdot \sigma_X ))$, $X = [x_1, x_2, \ldots, x_n]$
\end{center}
\vspace{-0.15cm}

S is the sigmoid function, $\mu_X$ and $\sigma_X$ are the mean and standard deviation of the vector $X$, $\alpha$ is an hyperparameter that controls the slope of the sigmoid and $z_k$ is a hyperparameter that controls the amount of sparsity. A higher $z_k$ value corresponds to more sparsity in the activations.
\vspace{-0.2cm}
\subsection{Hard ASH}
\vspace{-0.2cm}

\citet{lan2023elephant} theorized that activation functions should be better suited for continual learning if their gradients are fairly sparse, i.e. the activation function should be flat in most places. To reduce the gradient flow I replaced the sigmoid function with a hard sigmoid \citep{courbariaux2016binaryconnect} and clip the first $x_i$ term to values between 0 and 2. See appendix \ref{hard-ash} for the exact Hard ASH formula.

\begin{figure}[h]
\begin{center}
\includegraphics[width=\textwidth]{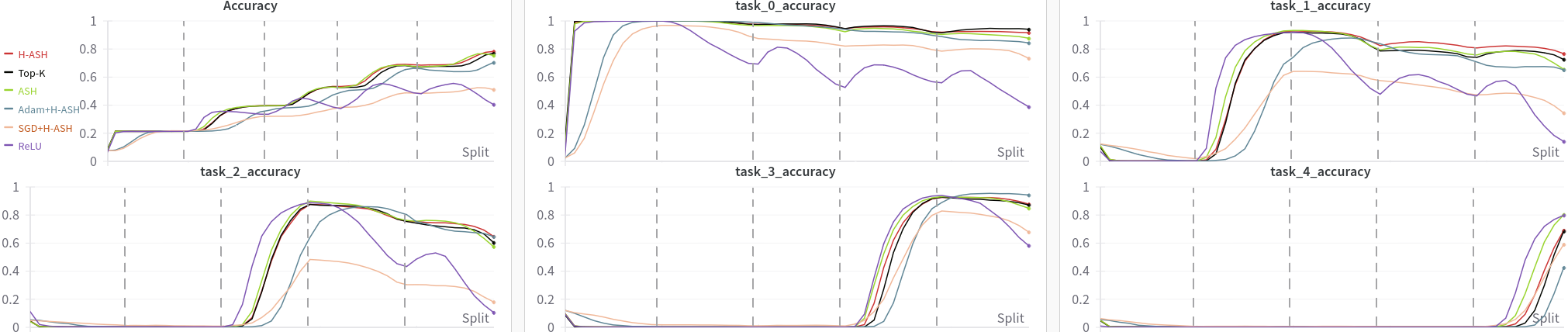}
\end{center}
\vspace{-0.5cm}
\caption{Overall and per-task validation accuracies of a single run of each method. Vertical lines represent the points in the training where the task changes. Optimizer is Adagrad when not specified. Best methods slowly lose accuracy on old tasks, but struggle to learn the last task. ReLU forgets the old tasks even with good optimizer like Adagrad. Meanwhile Hard ASH keeps some old-task performance even with plain SGD. Variations between runs are small enough to be barely visible.}
\vspace{-0.4cm}
\end{figure}
\vspace{-0.1cm}
\section{Experiment}
\vspace{-0.1cm}

I ran an experiment on 5 task Split-MNIST dataset in the class incremental setting, the hardest setting of the Split-MNIST, where a single network has to learn all the tasks without task id input \citep{hsu2019reevaluating}. To create the 5 tasks, the MNIST digits are split into 5 sets of 2 classes each and tasks are trained one after another with no replay of previous inputs. The network architecture was a simple MLP with one hidden layer of 1000 neurons, as in \citet{bricken2023sparse}. I trained each network for only 1 epoch to save compute.

The goal of the experiment was to test the effectiveness of various sparse activation functions with standard optimizers tuned for continual learning.
For each (activation function, optimizer) pair tested, I ran a hyperparameter sweep to find the best final mean accuracy. Before running the sweep, I manually estimated the best hyperparameters, listed in \ref{hyperparams}. Shown in table \ref{activation-functions-best-percent} are the best result for the main activation functions of the study. For full table of results see \ref{full-results}. 

All of sparse activation functions performed better than all of the non-sparse functions. Hard ASH was the best in almost every optimizer setup, followed by Top-K and ASH. Adagrad performed the best out of the optimizers tested, followed by RMSprop\citep{RMSProp} and Adam\cite{Adam}. I also ran SGD and SGDM for comparison, but both of these had lower accuracy than the optimizers with adaptive, per parameter, learning rates.
 
\vspace{-0.3cm}
\begin{table}[h]
\caption{
Activation functions with best performance across tested optimizers.
Average of 5 runs and 95\% C.I.
EWC\citep{Kirkpatrick_2017}, FlyModel\citep{shen2021algorithmic} and SDMLP+EWC baseline results are from \citet{bricken2023sparse} and also use an MLP with a single 1000 neuron hidden layer.
}
\label{activation-functions-best-percent}
\begin{center}
\begin{tabular}{lllll}
\multicolumn{1}{c}{\bf Activation / Method} & \multicolumn{1}{c}{\bf Epochs} & \multicolumn{1}{c}{\bf Mean accuracy} & \multicolumn{1}{c}{\bf Best optimizer} \\
\hline \\[-0.2cm]
ASH        & 1   & 76.4\% (±1.4\%)          & Adagrad \\
Hard ASH   & 1   & {\bf 78.3\%} (±1.4\%)    & Adagrad \\
Top-K      & 1   & 76.0\% (±1.6\%)          & Adagrad \\
ReLU       & 1   & 49.2\% (±7.9\%)          & Adam \\
\hline \\[-0.3cm]
EWC        & 500 & 61\%                     & SGD \\
SDMLP      & 500 & 69\%                     & SGD \\
SDMLP+EWC  & 500 & {\bf 83\%}               & SGDM \\
FlyModel   & 1   & 77\%                     & Association rule learning \\
\end{tabular}\end{center}
\end{table}
\vspace{-0.7cm}
\section{Conclusions}
\vspace{-0.2cm}
This study challenges the conventional approaches to continual learning by demonstrating that the we can get decent results in Split-MNIST even without any continual learning algorithms or task-related information. Hard ASH performed very well in my experiments. I suggest trying it as a faster to compute alternative to Top-K, that might also boost accuracy.

Choosing the best optimizer is a hard problem given the amount of options to choose from and the amount of hyperparameter tuning required for good performance. \citet{schmidt2021descending} alone listed over 100 known algorithms, each of which can optionally be paired with various learning rate schedules. For now, I recommend Adagrad as a relatively easy to tune default for continual learning. Appendices \ref{adam-bias-correction} and \ref{lr-schedule} provide more insight into optimizer performance.

\section{URM Statement}
The author acknowledges that at the author of this work meets the URM criteria of ICLR 2024 Tiny Papers Track.

\bibliography{iclr2023_conference_tinypaper}
\bibliographystyle{iclr2023_conference_tinypaper}

\appendix
\section{Appendix}

\subsection{Reproducibility}
The code to reproduce the experiments in this paper is publicly available here:
\begin{center}
   \url{https://github.com/LesserScholar/hard-ash}
\end{center}
Care has been taken to make sure the results are reproducible, including consistent use of JAX keys and deterministic shuffling of data.

\subsection{Hard ASH formula}
\label{hard-ash}

The formula I used for Hard ASH is:
\begin{center}
\
$HardSigmoid(x) = \dfrac{clip(x + 3, 0, 6)}{6}$
$Hard ASH(x_i) = clip(x_i, 0, x_{max}) \cdot  HardSigmoid(\alpha \cdot ( x_i - \mu_X - z_k \cdot \sigma_X))$
\end{center}

 Where $x_{max}$ is a hyperparameter that I always set to 2. The formula for hard sigmoid is the one used in in JAX  (jax.nn.hard\_sigmoid).

 Together the clip, the hard sigmoid and a high enough value for $\alpha$ cause most of the activations to be saturated (at either 0 or 2) and reduce amount of flowing gradients. Intuitively, this means that for each training example, I only update the incoming and outgoing weights for the activations where the network is unsure if that particular $x_i$ should be on or off.

\subsection{Network Initialization}
In all of my tests I used the standard Kaiming initialization \citep{he2015delving} for both layers of the MLP. It is likely that there are more efficient ways to initialize a Hard ASH network that account for the sparsity in the activations, but those explorations were not included in this study.

\subsection{Weight normalization}
I used weight normalization \citep{salimans2016weight} with a fixed $g$ of 1, only on the first layer of the MLP. In preliminary testing, weight normalization on the first layer consistently increased performance of multiple methods by 1 to 2 percentage points. Weight normalization on the second layer was, in preliminary testing, either net neutral or slightly negative. 

\subsection{Full results}
\label{full-results}

Tables \ref{optimizers-best-activation-percent} and \ref{all-results-table} show how all the methods and method pairs stacked against one another.

It is noteworthy that all the sparse activation functions kept a decent chunk of their performance even with basic SGD.

In the full results I tried two different versions of Top-K, subtract and mask. The difference between the two methods is that in Top-K subtract, the \textit{k}-th highest value is subtracted from the activations before masking. In the main text only Top-K subtract is used, since it performs better. \citet{bricken2023sparse} also found Top-K subtract to perform better than Top-K mask.

I also tested LWTA\citep{srivastava_2013, xiao2019enhancing} which has been suggested as a faster to compute alternative to Top-K, but found it's performance to be worse than Top-K subtract or ASH based functions.

\begin{table}[h]
\caption{Best results for each optimizer. Average of 5 runs and 95\% C.I.}
\label{optimizers-best-activation-percent}
\begin{center}
\begin{tabular}{lll}
\multicolumn{1}{c}{\bf Optimizer}  &\multicolumn{1}{c}{\bf Mean Accuracy} &\multicolumn{1}{c}{\bf Best Activation Function}
\\ \hline \\
Adagrad     & {\bf 78.3\%} (±1.4\%) & Hard ASH \\
RMSprop     & 77.7\% (±1.8\%) & Hard ASH \\
Adam        & 71.6\% (±1.6\%) & Hard ASH \\
SGDM        & 66.3\% (±3.4\%) & Top-K Subtract \\
SGD         & 52.9\% (±5.3\%) & Hard ASH \\
\end{tabular}
\end{center}
\end{table}

\begin{longtable}{|l|c|c|c|c|c|}
\caption{Full results for all tested activation functions with all tested optimizers. Average of 5 runs and 95\% C.I. Poorly performing combinations were terminated early to save compute and thus have larger error bounds.}
\label{all-results-table} \\
\hline
Activation & Adagrad & RMSprop & Adam & SGDM & SGD \\
\hline
\endfirsthead

\multicolumn{6}{c}{{Table \thetable\ Continued from previous page}} \\
\hline
Activation & Adagrad & RMSprop & Adam & SGDM & SGD \\
\hline
\endhead

\hline
\multicolumn{6}{|r|}{{Continued on next page}} \\
\hline
\endfoot

\hline
\endlastfoot

ASH             & 76.4 ± 1.4 & 75.7 ± 1.2 & 69.5 ± 2.0 & 65.0 ± 0.4 & 52.4 ± 8.6 \\
Hard ASH        & 78.3 ± 1.4 & 77.7 ± 1.8 & 71.6 ± 1.6 & 65.1 ± 1.6 & 52.9 ± 7.6 \\
Top-K subtract  & 76.0 ± 1.6 & 75.0 ± 2.9 & 71.5 ± 1.4 & 66.3 ± 3.5 & 51.5 ± 10.4 \\
Top-K mask      & 65.0 ± 4.6 & 69.7 ± 0.8 & 67.9 ± 2.4 & 62.9 ± 3.9 & 44.1 ± 15.3 \\
LWTA            & 67.2 ± 2.6 & 67.1 ± 2.0 & 64.9 ± 2.2 & 61.3 ± 4.1 & 39.9 ± 14.4 \\
ReLU            & 43.8 ± 10.7 & 39.7 ± 11.7 & 49.2 ± 9.7 & 35.7 ± 2.2 & 19.8 ± 4.0 \\
SwisH           & 46.2 ± 9.7 & 41.4 ± 10.1 & 49.2 ± 9.7 & 51.9 ± 2.8 & 26.5 ± 8.3 \\
Sigmoid         & 52.4 ± 10.3 & 44.2 ± 9.7 & 35.3 ± 8.2 & 20.8 ± 5.0 & 15.6 ± 1.0 \\
Hard Sigmoid    & 56.5 ± 1.2 & 48.1 ± 11.6 & 32.0 ± 9.2 & 19.4 ± 1.4 & 14.7 ± 0.8 \\

\end{longtable}

\subsection{Hyperparameters}
\label{hyperparams}

Tables \ref{common-config} and \ref{method-specific-hyperparameters} list the hyperparameters for each method used in the study. 

The optimizer parameters are heavily tuned towards better performance in Split-MNIST. For momentum optimizers the momentum values were set higher than usual. For RMSprop the decay is set very high. For Adagrad the initial fill value is set unusually low which makes it's behavior similar to RMSprop in the beginning of the training, i.e. the learning rates are very high at the start of the training.

For Top-K the best $k$ values were 64 and 96, corresponding to sparsity between 94\% and 90\%. For LWTA the best amounts of groups were 25 and 50, 97.5\% and 95\% sparse, respectively. \citet{lee2022stochastic} shows how to calculate density of ASH based on value of $z_k$. I found that a $z_k$ value of 2.0-2.5 yielded the best results, which corresponds roughly to 97-99\% sparsity.

\begin{table}[h]
\caption{Common hyperparameters}
\label{common-config}
\begin{center}
\begin{tabular}{ll}
\multicolumn{1}{c}{\bf Hyperparameter}  &\multicolumn{1}{c}{\bf Values}
\\ \hline \\
Batch size & 64 \\
First layer weight norm & True \\ 
Second layer weight norm & False \\
Hidden size       & {1000} \\
Gradient clip & 0.01 \\
\end{tabular}
\end{center}
\end{table}
\begin{table}[ht]
\caption{Method Specific Hyperparameters}
\label{method-specific-hyperparameters}
\begin{center}
\begin{tabular}{lll}
\multicolumn{1}{c}{\bf Method}  &\multicolumn{1}{c}{\bf Hyperparameter} &\multicolumn{1}{c}{\bf Values}
\\ \hline \\
Activations & & \\
\hline
Ash             & $\alpha$       & {3.0, 4.0} \\
                & $Z_k$          & {2.2, 2.3, 2.4} \\
                & Hard ASH $x_{max}$ & 2.0 \\
\hline
Top-K           & k              & {32, 64, 96, 128, 256} \\
\hline
LWTA            & groups         & {25, 50, 100} \\
\hline
Optimizers & & \\
\hline
RMSprop         & decay          & {0.998, 0.999, 0.9991, 0.9992, 0.9993} \\
                & learning rate  & {4e-6, 5e-6, 5.5e-6, 6e-6, 8e-6} \\
\hline
Adam            & $\beta_1$      & {0.9, 0.95, 0.98, 0.99} \\
                & $\beta_2$      & {0.999, 0.9995} \\
                & learning rate  & {8e-6, 1e-5, 1.5e-5} \\
\hline
Adagrad         & learning rate  & {1e-4, 2e-4, 3e-4} \\
                & initial value  & 1e-6               \\
\hline
SGD             & learning rate  & {3e-4, 4e-4, 5e-4} \\
\hline
SGDM            & momentum       & {0.99, 0.992, 0.994, 0.996} \\
                & learning rate  & {8e-6, 1e-5, 1.5e-5} \\
\end{tabular}
\end{center}
\end{table}

\pagebreak

\subsection{Performance without task splits}

The continual learning optimized hyperparameters used in the main experiment sweep have been tuned only for Split-MNIST task. Using such high adaptive learning rate and momentum parameters is very unusual. To assess how much performance is lost using these settings in typical MNIST without the task splits, I ran through the same sweep of activation functions and optimizers but training all classes simultaneously, i.e. with i.i.d. data.

Table \ref{iid-perf} shows the results when trained on whole MNIST at once (i.e. with i.i.d. dataset). First sweep was with the same hyperparameters used in the main experiment, tuned for continual learning, and second with the optimizer hyperparameters tuned to i.i.d data. In the second case, the much lower momentum and higher learning rates, allow the model to reach much better accuracy in 1 epoch.

\begin{table}[h]
\caption{
Comparison of split task continual learning scenario, i.i.d. with continual learning optimizer and i.i.d with regular optimizer.
Average of 5 runs and 95\% C.I.
}
\label{iid-perf}
\begin{center}
\begin{tabular}{llll}
\multicolumn{1}{c}{\bf Method} &\multicolumn{1}{c}{\bf Epochs}  &\multicolumn{1}{c}{\bf Mean accuracy}
\\ \hline \\
Hard ASH /w task splits and Split-MNIST optimizer  & 1   & 78.3\% (±1.4\%)\\
ReLU i.i.d. dataset and Split-MNIST optimizer & 1   & 91.3\% (±2.5\%) \\
ReLU i.i.d. dataset and normal optimizer & 1   & 96.9\% (±3.3\%) \\
\end{tabular}
\end{center}
\end{table}

\subsection{Adam and bias correction}
\label{adam-bias-correction}
In the main text I focused on evaluating existing and well studied optimizers without modifications. One of the more surprising results was how much worse Adam performed when compared to very similar RMSprop and Adagrad algorithms. After testing I found that I can boost the Adam performance almost to RMSprop level simply by removing bias correction from the algorithm.

Bias correction in Adam adjusts the first and second moment estimates to account for their initialization at zero. This makes the moment estimations more accurate (i.e. unbiased) during the early optimization steps.\citep{Adam} Unlike Adam, standard implementations of RMSprop do not use bias correction. RMSprop and biased Adam performing significantly better than standard Adam suggests to us that the initial very high learning rates given by the biased exponential averaging are important for the final accuracy of the network.

\begin{table}[h]
\caption{Testing Adam performance with bias correction removed. Average of 5 runs and 95\% C.I.}
\label{adam-bias-table}
\begin{center}
\begin{tabular}{lll}
\multicolumn{1}{c}{\bf Optimizer}  &\multicolumn{1}{c}{\bf Mean Accuracy} &\multicolumn{1}{c}{\bf Activation Function}
\\ \hline \\
Adam with bias correction        & 71.6\% (±1.5\%) & Hard ASH \\
Adam without bias correction        & 76.7\% (±1.5\%) & Hard ASH \\
\end{tabular}
\end{center}
\end{table}

\subsection{Learning rate schedules}
\label{lr-schedule}
In my experiment adaptive learning rate methods did very well, but I did not test learning rate schedules. Continual learning experiments are often done with constant learning rates, to maintain plasticity (for more on plasticity see \ref{permuted-mnist}). But the success of Adagrad, RMSprop and biased Adam suggests that the initial high learning rate is important. Therefore I also tested SGD with exponential decay schedule and found it to be fairly effective. I got accuracy of {\bf 72.4\%} (±3.2\%), combining Hard ASH with exponentially decaying learning rate with decay of 0.7 every 200 steps, with starting learning rate of $3.35e-3$. The parameters for exponentially decaying learning rate are fairly sensitive and much harder to tune than something like Adagrad with a constant learning rate.

\subsection{Plasticity experiment on permuted MNIST}
\label{permuted-mnist}
Of primary concern in continual learning is the so called stability-plasticity dilemma \citet{stability_plasticity}, which highlights the tension between remembering past tasks and learning to perform on new tasks. As an extreme example, it would be easy to build a learning system that perfectly remembers the early tasks and then stops learning, simply by setting the learning rate close to 0 after certain amount of training steps. But this would be against the spirit of continual learning since the network would completely stop absorbing new knowledge. So the end goal of continual learning is to build a learning system where the performance on old tasks is stable and it keeps learning new tasks easily.

To assess the effect that sparsity has on plasticity, I ran another smaller experiment on the permuted MNIST dataset \citep{Kirkpatrick_2017}. In permuted MNIST, each subsequent task in a sequence of tasks is created by applying a fixed permutation to the pixels of the original MNIST images, resulting in different, but structurally similar, tasks. Since you can keep applying new permutations at will, you can keep training on new tasks almost forever.

To test the effectiveness of sparsity on plasticity, I ran this experiment on Hard ASH with different hyperparameters. The hypothesis tested was that changing the amount of sparsity (modifying $z_k$) or gradient sparsity (modifying $\alpha$, steeper alpha curve results in more sparsity in the gradients) would have an effect on the plasticity of the network.

\begin{figure}[h]
\begin{center}
\includegraphics[width=13cm]{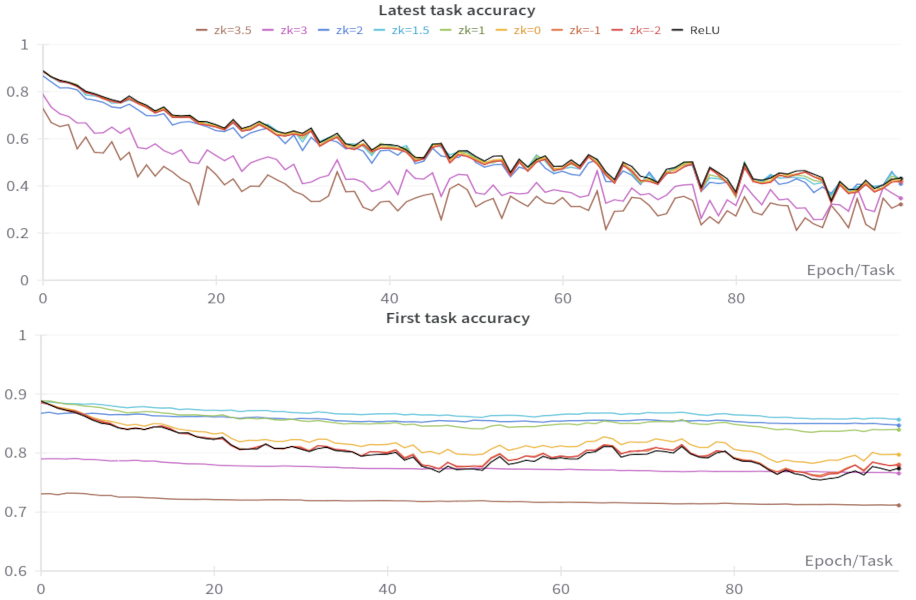}
\end{center}
\caption{Latest task and first task validation accuracies when varying $z_k$}
\label{vary_zk}
\end{figure}

The experiment was ran for 100 epochs, changing to a new task after every epoch for a total of 100 tasks. Adagrad was chosen as the optimizer since it was the overall best performer in the main study. Network architecture is the same as in the main experiment, an MLP with a single 1000 neuron hidden layer.

\begin{figure}[h]
\begin{center}
\includegraphics[width=13cm]{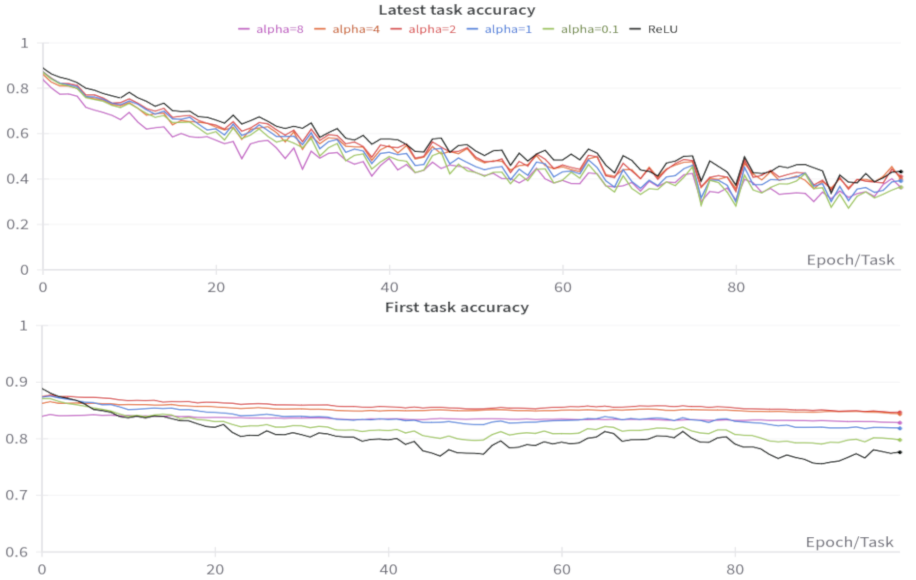}
\end{center}
\caption{Latest task and first task validation accuracies when varying $\alpha$}
\label{vary_alpha}
\end{figure}

Figures \ref{vary_zk} and \ref{vary_alpha} show the accuracy on the latest task right after it's training is complete (plasticity) and performance on the very first task through the whole training (stability).

In figure \ref{vary_zk}, I show the effect of varying $z_k$, i.e. the amount of sparsity in the hidden layer representation. Latest task accuracy graph shows that varying the amount of sparsity has negligible effect on plasticity. When $z_k < 3$, the Hard ASH MLP has pretty much same plasticity as baseline ReLU MLP. When $z_k \geq 3$, the performance is already worse during the first epoch, but it doesn't appear to help much with plasticity.

In the first task accuracy graph we see that the networks ability to retain accuracy on the first task increases with the amount of sparsity. When $z_k < 1$, the network performs similarly to the baseline ReLU, but when $z_k > 1$, the performance on the first task falls only slightly during the training of the 99 subsequent tasks.

\vspace{-0.1cm}
\vspace{-0.1cm}

In figure \ref{vary_alpha}, I show the effect of varying $\alpha$ or the steepness of the activation slope, i.e. the amount of sparsity in the gradients. For this experiment, $z_k$ was held constant at 1.5. Latest task accuracy graph shows much of the same as figure \ref{vary_zk}. Plasticity is the same regardless of the settings and pretty much matches the baseline ReLU. The stability, or the ability to retain accuracy on the very first task, increases with $\alpha$. When $\alpha$ gets very high (e.g. 8, as shown in the graph) there is a noticeable performance degradation already at the beginning of the training, but no significant benefits to plasticity.

Thus my conclusion is that representation sparsity alone is not enough to solve plasticity in continual learning and something extra is required. Instead I can say that sparsity helps with stability without significant penalty to plasticity. For a more complete continual learner, that overcomes both sides of stability-plasticity dilemma, we could try to combine sparse representations with a technique that increases plasticity, such as Continual backprop \citep{dohare2022continual}, Shrink and perturb \citep{ash2020warmstarting} or progressive magnitude based pruning with model expansion like in \citet{menick2020practical}.

\end{document}